\pdfoutput=1

\documentclass[11pt]{article}

\usepackage[]{acl}

\usepackage{times}
\usepackage{latexsym}

\usepackage[T1]{fontenc}

\usepackage[utf8]{inputenc}

\usepackage{microtype}

\usepackage{amsfonts}
\usepackage{amsmath}
\usepackage{amssymb}
\usepackage{graphicx}
\usepackage{multirow}
\usepackage{caption}
\usepackage{subfigure}
\usepackage{enumitem}
\usepackage{float}
\usepackage{makecell}
\usepackage{booktabs}
\usepackage{tabu}
\usepackage{tabularx}
\usepackage{pifont}

\title{CALCS 2021 Shared Task: Machine Translation for Code-Switched Data}

\author{
Shuguang Chen \textsuperscript{1},
Gustavo Aguilar \textsuperscript{1},
Anirudh Srinivasan \textsuperscript{2},
Mona Diab \textsuperscript{3} \and 
Thamar Solorio \textsuperscript{1} \\

University of Houston \textsuperscript{1} \\
The University of Texas at Austin \textsuperscript{2}\\
Meta AI \textsuperscript{3} \\

\{schen52, gaguilaralas, tsolorio\}@uh.edu \textsuperscript{1} \\ 
anirudhsriniv@gmail.com \textsuperscript{2}\\
mdiab@fb.com \textsuperscript{3}
}

\begin{document}
\maketitle
\begin{abstract}
To date, efforts in the code-switching literature have focused for the most part on language identification, POS, NER, and syntactic parsing. In this paper, we address machine translation for code-switched social media data. We create a community shared task. We provide two modalities for participation: supervised and unsupervised. For the supervised setting, participants are challenged to translate English into Hindi-English (Eng-Hinglish) in a single direction. For the unsupervised setting, we provide the following language pairs: English and Spanish-English (Eng-Spanglish), and English and Modern Standard Arabic-Egyptian Arabic (Eng-MSAEA) in both directions. We share insights and challenges in curating the "into" code-switching language evaluation data. Further, we provide baselines for all language pairs in the shared task. The leaderboard for the shared task comprises 12 individual  system submissions corresponding to 5 different teams. The best performance achieved is 12.67\% BLEU score for English to Hinglish and 25.72\% BLEU score for MSAEA to English.
\end{abstract}

\section{Introduction}
\label{sec: introduction}

Linguistic code-switching refers to the linguistic phenomenon of alternating between two or more languages or varieties of language both across sentences and within sentences (aka intra sentential code-switching, which is the type we focus on in this work)\citep{joshi-1982-processing}. The wide use of social media platforms (e.g, Twitter, Facebook, Reddit, etc.) where users communicate with each other more spontaneously, rendering significantly more written code-switched data. Accordingly, this raises an increasing demand for more tools and resources to process code-switched data. However, current NLP technology is lagging in the development of resources and methodologies that can effectively process such phenomena. This is true for even the large multilingual pre-trained models such as mBART \citep{liu-etal-2020-multilingual-denoising} and mT5 \citep{xue-etal-2021-mt5}. At the same time, the growing adoption of smart devices and automated assistants that rely on speech interfaces makes it even more pressing for the NLP field to deal with code-switched language data. 

\begin{figure}[t]
\centering
\subfigure{
    \begin{minipage}[t]{\linewidth}
    \centering
    \includegraphics[width=1\linewidth]{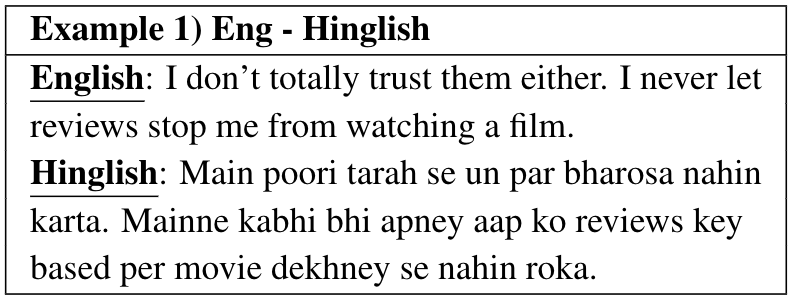}
    \end{minipage}
}
\subfigure{
    \begin{minipage}[t]{\linewidth}
    \centering
    \includegraphics[width=1\linewidth]{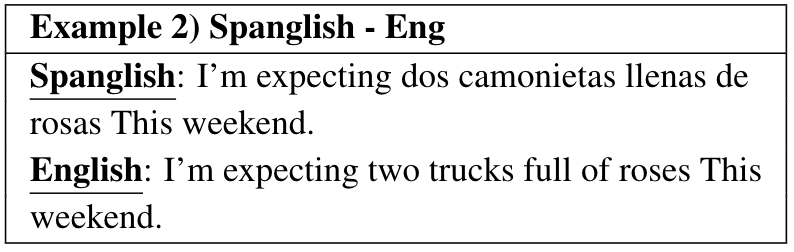}
    \end{minipage}
}
\caption{Examples from the dataset. The irregular grammar and non-standard syntactic rules make it difficult to translate from one language to the other. }
\label{fig: examples}
\vspace{-0.5cm}
\end{figure}
In this work, we present the new task of machine translation (MT) from and to Code Switched language. We create a community task, defining and curating datasets and evaluation metrics. We leverage already existing standard code-switching datasets that were created in the community as used in the previous workshops on Computational Approaches to Linguistic Code-Switching (CALCS). We created a series of shared tasks focusing primarily on enabling technology for code-switching, including language identification (LID) \citep{solorio-etal-2014-overview}, part of speech (POS) tagging \citep{molina-etal-2016-overview}, and named entity recognition (NER) \citep{aguilar-etal-2018-named}. In our task, we leverage previously created data sets and extend them for the task of machine translation (MT) under code-switching settings in multiple language combinations and directions. General challenges to the processing of code-switched language persist in this setting but notably, the following challenges are especially significant to the MT problem space: 1) irregular grammar and spelling of utterances; 2) non-standard syntactic rules; and 3) ambiguous words (e.g, `car` in French and English), etc. For example, in Figure \ref{fig: examples}, the irregular grammar and non-standard syntactic rules make it difficult to translate from one language to the other. With this in mind, we host a new community task for machine translation on code-switched data. The goal is to provide code-switched MT datasets and to continue to motivate new research and energize the NLP community to take on the challenges posed by code-switched data. 

For the community task, we provide two modes for participation: supervised and unsupervised. For the supervised setting, we ask participants to translate English into Hindi-English (Eng-Hinglish) in a single direction. For the unsupervised setting, we provide the following language pairs: English and Spanish-English (Eng-Spanglish), and English and Modern Standard Arabic-Egyptian Arabic (Eng-MSAEA) in both directions. We received 12 individual public system submissions from 5 different teams. The best performance achieved is 12.67\% BLEU score for English to Hinglish and 25.72\% BLEU score for MSAEA to English.
Our contributions are multi-fold:
\begin{enumerate}
    \item We introduce the novel task of translating into code-switched language;
    \item We create new standard datasets for evaluating MT for code-switched data: English$\rightarrow$Hinglish, English-Modern Standard Arabic-Egyptian Arabic, and English-Spanglish;
    \item We provide baseline systems for MT for several pairs of languages: English-Hinglish, English-Spanglish; 
    \item We present and discuss some of the challenges of generating evaluation data for code-switched language.
\end{enumerate}

\section{Related Work}
Code-switching is a phenomenon where multilingual speakers communicate by switching back and forth between the languages they speak or write when communicating with other multilingual speakers \citep{joshi-1982-processing}. It tends to be more common in informal settings such as social media platforms where interactions are more casual. Previous work has studied code-switching for many languages from different perspectives. \citet{Toribio2001AccessingBC} presents new methodologies for gathering code-switched data, examining permissible and unacceptable language alternations. \citet{goyal2003bilingual} introduces a bilingual syntactic parser that operates on input strings from Hindi and English and develops grammars for Hindi and English. Lately, researchers mainly focus on exploring this phenomenon at the core level of NLP pipelines, including language identification \citep{solorio-liu-2008-learning, aguilar-solorio-2020-english}, language modeling \citep{garg-etal-2018-code, gonen-goldberg-2019-language}, and Named Entity Recognition \citep{winata-etal-2019-learning, aguilar-etal-2018-named}, etc. 

Recent advances in large pre-trained models \citep{devlin-etal-2019-bert, lewis-etal-2020-bart} have led to significant performance gains in many multilingual tasks. Without further pre-training and extra neural network components, these large pre-trained models (e.g, mBART \citep{liu-etal-2020-multilingual-denoising} and mT5 \citep{xue-etal-2021-mt5}) can be easily adapted to code-switching tasks. However, the pre-training tasks of these models are performed on monolingual text without language alternation. Applying these techniques to the task of code-switching machine translation may not result in a good performance due to the challenges mentioned in Section \ref{sec: introduction}.

Machine translation on code-switched data has been recently receiving increasing attention in the NLP community. For this task, there are two main focused efforts: 1) data resources, and 2) efficient methods. For data resources, \citet{srivastava-singh-2020-phinc} presents a parallel corpus for Hindi-English (Hinglish) and English and proposes a translation pipeline built on top of Google Translate. \citet{tarunesh-etal-2021-machine} adapts a neural machine translation model to automatically generate code-switching Hindi-English data. However, due to the lack of resources, this phenomenon remains challenging, especially for social media genres where annotated data is limited. In this work, in addition to the Hinglish-English dataset, we further propose standardized datasets for translating from monolingual data to code-switched data. Critically, we are proposing standard datasets for evaluating to and from code-switched data. For efficient methods, \citet{singh2017towards} employs the n-best-list generation approach to address the spelling variation in the text and also enrich the resource for translation. \citet{yang-etal-2020-csp} proposes a simple yet effective pre-training approach to align the cross-lingual information between the source and target language. \citet{xu-yvon-2021-traducir} designs a new method to simultaneously disentangle and translate the two mixed languages. In this paper, we compare different systems from the participating teams in terms of methodology (e.g, model architecture and training strategy) and performance. We further provide in-depth analysis. In the spirit of reproducibility, we make our data and experiments publicly available to support future research in this area.

\section{Task Description}
The goal of the community task is to motivate new research and energize the community to take on challenges posed by code-switched data. Most research to date has focused on translation from code-switched language to monolingual language, albeit without establishing standard data sets for the task. In our work, we establish standard datasets in three different language pairs. Moreover, we introduce the novel task of translating into code-switched language. Generating \textit{plausible} code-switched language is a challenge for systems in general and it is still an untapped novel territory in NLP. As mentioned earlier, code-switching is pervasive among multilingual speakers as well as among even monolingual speakers given the global nature of communication. Catering to such users requires systems that are able to produce more natural language consistent and coherent with their manner of speaking, e.g. in conversational AI. We contend that such natural language interactions with code-switching understanding and generation will be the next frontier in human language technology interaction. Hence an initial challenge is to create standardized evaluation datasets that reflect the code-switched phenomenon. Machine translation serves both ends exemplifying understanding and generation. With this in mind, we hosted a new shared community task on machine translation for code-switched language. The shared task mainly focuses on translating a source sentence into a target language while one of the directions contains an alternation between two languages.\footnote{We constrain the problem space to code-switching being on one side of the directions for this initial iteration of the task, however, we plan on extending the datasets to include code-switched data source to code-switched data target in the near future.} Given a monolingual/code-switched sentence, the goal is to translate it into a code-switched/monolingual sentence, respectively. In this shared task, we provide the following two settings for participation:

\paragraph{Supervised settings}
We primarily focus on translating from monolingual text to code-switched text. This setting only has one language pair with a single direction, i.e, English$\rightarrow$Hinglish. We provide parallel training data and participants are required to build systems for translating an English sentence to a Hinglish sentence. 

\paragraph{Unsupervised settings}
This setting comprises  two language pairs: English to/from Spanish-English (Eng-Spanglish), and English to/from Modern Standard Arabic-Egyptian Arabic (Eng-MSAEA). We focus on machine translation in two directions, either from monolingual text to code-switched text or from code-switched text to monolingual text. To this end, we provide raw data with no reference translations (no parallel data). Participants are required to provide translations for both directions for either or both language pairs.

\paragraph{}
Additionally, in both settings, participants are allowed to use any external resources. We employ the BLEU score metric for evaluation as proposed by \citet{post-2018-call}. Participants can submit their predictions for test data to the Linguistic Code-switching Evaluation (LinCE) platform \citep{aguilar-etal-2020-lince}. The leaderboard in the LinCE platform averages the BLEU scores across multiple language pairs and directions to determine the position. Participants can submit their results while comparing with others in real-time.

\section{Datasets}
\begin{table*}[ht]
\small
    \centering
    \renewcommand{\arraystretch}{1.2}
    \resizebox{\linewidth}{!}{
        \begin{tabular}{l|lll|ll|ll|ll|ll}
        \hline
            & \multicolumn{3}{c|}{\bf Supervised} & \multicolumn{8}{c}{\bf Unsupervised} \\
        \cmidrule(lr){2-12}
            & \multicolumn{3}{c|}{\bf Eng$\rightarrow$Hinglish} & \multicolumn{2}{c|}{\bf Eng$\rightarrow$Spanglish} & \multicolumn{2}{c|}{\bf Spanglish$\rightarrow$Eng} & \multicolumn{2}{c|}{\bf Eng$\rightarrow$MSAEA} & \multicolumn{2}{c}{\bf MSAEA$\rightarrow$Eng}\\
        \cmidrule(lr){2-12}
            & \bf Train & \bf Dev & \bf Test & \bf Raw & \bf Test & \bf Raw & \bf Test & \bf Raw & \bf Test & \bf Raw & \bf Test\\
        \hline
        Sentences       & 8,060     & 960       & 942       & 15,000    & 5,000     & 15,000        & 6,500         & 15,000    & 5,000   & 12,000         &6,500 \\
        \hline
        Tokens              & 93,025    & 11,465    & 11,849    & 166,649   & 212,793   & 228,484   & 189,465   & 166,649    & 104,429   & 219,796  & 303,998 \\
        Eng tokens          & 33,886    & 4,273     & 4,333     & 116,617   & 97,646    & 43,869    & 100,000   & 116,617    & 38,967    & -        & 181,890 \\
        Non-Eng tokens      & 58,381    & 7,130     & 7,420     & -         & 62,587    & 98,412    & 41,839    & -         & 36,676    & 149,087   & 80,780 \\
        Other tokens        & 758       & 62        & 96        & 50,032    & 52,560    & 86,203    & 47,626    & 50,032    & 28,831    & 70,709    & 41,328 \\
        \hline
        \end{tabular}
    }
    \caption{Distribution Statistics for the data sets for each language pair. Only English-Hinglish had training and development data. English-Spanglish and English-MSAEA both had test data with associated gold translations, as well as raw data. For the raw sets, we just provide the raw data in one of directions without gold translations. \textbf{Other tokens} refer to social media special tokens, e.g, username mentions, emoji expressions and urls.
    }
    \label{tab: data_statistics}
\end{table*}

\subsection{Eng-Hinglish}
\paragraph{Data Curation}
The CMU Document Grounded Conversations Dataset \cite{zhou-etal-2018-dataset} contains a set of conversations between users, with each conversation being grounded in a Wikipedia article about a particular movie. A subset of this dataset is translated into Hinglish (code-switched Hindi-English) and is available as the CMU Hinglish Document Grounded Conversations Dataset \cite{Gupta_Potluri_Reganti_Lara_Black_2019}. The English sentences from the former dataset and their corresponding Hinglish translations make up the parallel corpus used for our current task. The dataset has multiple translations for some of the English sentences. These sentences are pre-processed to remove newline and tab characters. These parallel sentences could be used for machine translation in either of the 2 directions (English $\rightarrow$ Hinglish or Hinglish $\rightarrow$ English). We chose the former as it involves the generation of code-switched text and represents a more challenging task overall. It is worth noting that this setup renders a single Hinglish translation per English sentence. 

\paragraph{Data distribution:}
The data statistics are shown in Table \ref{tab: data_statistics}. We use the same train/dev/test splits as originally created by \citet{zhou-etal-2018-dataset}. 

\subsection{Eng-Spanglish}
\paragraph{Data Curation}
For the Eng-Spanglish language pair, we leverage the English-Spanish
language identification corpus introduced in the first CALCS shared task \citet{solorio-etal-2014-overview} and the second CALCS shared task \citet{molina-etal-2016-overview}. We randomly sample a subset from this dataset where we add translations for both directions, namely the task for this language pair comprises two directions: \textit{Eng $\rightarrow$ Spanglish} and \textit{Spanglish $\rightarrow$ Eng}. As the original data was collected from Twitter, we normalized some social media special tokens, e.g, replace username mentions with `<username>'. Since this data is used in the unsupervised setting, we do not provide training data. We only provide test data. For the latter data, we curate reference translations for both directions from three translators. 

\paragraph{Data distribution}
The data statistics for both  directions \textit{Eng $\rightarrow$ Spanglish} and \textit{Spanglish $\rightarrow$ Eng} are shown in Table \ref{tab: data_statistics}. We also provide raw data from the code-switched Spanglish data from previous collections. Those statistics are also shown in Table \ref{tab: data_statistics}. The idea for providing the raw data is to provide the participants with a sense of the data genres and domains and the code-switching style of the data used in the task. As shown in Table \ref{tab: data_statistics}, it is worth noting that there are no Spanish tokens in the raw data for \textit{Eng $\rightarrow$ Spanish}. Likewise, the ratios of English tokens for \textit{Spanglish $\rightarrow$ Eng} in raw and test data are 19.2\% and 19.5\%. The skewed distribution poses a great challenge considering that the model may not see enough words in the target language. However, we think that the skewness can be reasonably handled with the provided data. Moreover, the raw and test sets draw a very similar data distribution, which can also help adapt the learning from training to testing.

\subsection{Eng-MSAEA}
\paragraph{Data Curation}
For the Eng-MSAEA language pair, we combine the datasets introduced in the CALCS-2016 \citep{molina-etal-2016-overview} and CALCS-2018 \citep{aguilar-etal-2018-named} as the new corpora for the current machine translation shared task. The data was collected from the Twitter platform and Blog commentaries. The tweets that have been deleted or belong to the user whose accounts have been suspended were removed and eliminated. We also perform normalization on each tweet to reduce the impact of social media special tokens. Similar to the Eng-Spanglish setting, we provide only test data and raw data. 

\paragraph{Data distribution}
We have two directions for Eng-MSAEA language pair:  \textit{Eng $\rightarrow$ MSAEA} and \textit{MSAEA $\rightarrow$ Eng}. The data statistics for each direction are listed in Table \ref{tab: data_statistics}. From Table \ref{tab: data_statistics}, we can see that there are no tokens from the target language in the raw data. 

\subsection{General Translation Guidelines}
The following are some of the overall guidelines used for curating translations: 
The translators are required to follow the standard grammar rules and typos should be corrected in the translation. The translated text should also be normalized. Slang, special forms, and expressions are to be translated to literal equivalents or to a more formal/standard form. The typos should be eliminated in  translation, e.g, ``Happo Birthday" should be translated to ``Feliz Cumpleaños". Additionally, speech effects should not be reflected in translation, e.g, ``Happy Birthdayyy" should be translated to  ``Feliz Cumpleaños". Social media special tokens, including username mentions, URLs, hashtags, and emojis, should not be translated even if they are transferable/translatable. Moreover,  measurement units should not be converted or localized, e.g, ``2LB" should not be translated to ``1kg". When translating from Spanglish into English, The translators were instructed to reword the English fragments when the translation warrants it. For  translation from code-switched text to monolingual text, we requested one gold reference. For the case where the target translation is in the code-switched language, we requested three gold references. It is worth noting that we did not prescribe how and when the translations should include a code-switch. We only emphasized that we expect the resulting language to be a plausible code-switching of the two languages. We also provided ample examples per language pair. We expect there to be significant variation among the translations, as different translators would choose to code-switch at different points. We hope that this elicited data set could be eventually compared to naturally occurring code-switched data to investigate where and how they vary.  

\subsection{Dataset Evaluation}
\begin{table*}[ht]
\small
    \centering
    \small
    \renewcommand{\arraystretch}{1.2}
    \resizebox{0.7\linewidth}{!}{
        \begin{tabular}{l|cccccc}
        \hline
        \multirow{2}{*}{Datasets}       & \multicolumn{6}{c}{\bf Code-switching Evaluation Metrics} \\
        \cmidrule(lr){2-7}
                                        & CMI & M-Index & I-Index & LE & SE & Burstiness \\
        \hline
        Eng$\rightarrow$Hinglish        & 36.81 &  0.90 &  0.53 &  0.95 &  2.13 & -0.07 \\
        Eng$\rightarrow$Spanglish       & 49.99 &  2.13 &  0.40 &  1.06 &  1.60 & -0.19 \\
        Spanglish$\rightarrow$Eng       & 48.98 &  2.19 &  0.39 &  1.06 &  1.57 & -0.15 \\
        Eng$\rightarrow$MSAEA           &  9.54 &  0.71 &  0.62 &  0.59 &  2.31 & -0.10 \\
        MSAEA$\rightarrow$Eng           & 10.07 &  0.52 &  0.76 &  0.58 &  3.72 & -0.03 \\
        \hline
        \end{tabular}
    }
    \caption{Code-switching statistics on each dataset. }
    \label{tab: data_evaluation}
\end{table*}
To measure the quality of the datasets we created, and understand the inherent characteristics of code-switched corpora, we evaluate our datasets following the statistics proposed by \citet{guzman2017metrics} and \citet{gamback2014measuring}, including:
\begin{itemize}
	\item \textit{Code-Mixing Index (CMI)}: The fraction of total words that belong to languages other than the most dominant language in the text;
	\item \textit{Multilingual Index (M-Index)}: A word-count-based measure that quantifies the inequality of the distribution of language tags in a corpus of at least two languages;
	\item \textit{Integration Index (I-Index)}:  A proportion of how many switch points exist relative to the number of language-dependent tokens in the corpus;
	\item \textit{Language Entropy (LE)}: the number of bits of information are needed to describe the distribution of language tags;
	\item \textit{Span Entropy (SE)}: the number of bits of information are needed to describe the distribution of  language spans;
	\item \textit{Burstiness}: A measure of quantifying whether switching occurs in bursts or has a more periodic characteristic.
\end{itemize}

The code-switching evaluation statistics of our datasets are shown in Table \ref{tab: data_evaluation}.

\section{Methods}
We received submissions from five different teams. Four teams submitted system responses for Eng-Hinglish, making this language pair the most popular in this shared task. On the other hand, we had no external submissions for Spanglish-Eng.

Below we provide a brief description of the two baselines as well as participant systems:

\begin{itemize}[]
    \item \textbf{baseline1} We use mBART \citep{liu-etal-2020-multilingual-denoising} as the baseline model trained for 5 epochs with a batch size of 8 and a learning rate of 5e-5. For the supervised setting, we simply fine-tune it on the parallel data. For the unsupervised setting, we directly generate translations without fine-tuning.
    \item \textbf{Echo (baseline)} This baseline simply passes inputs as outputs and the goal is to measure how much the overlap in input/output can contribute to the final performance. It is inspired by preliminary observations that there is a high token overlap between the source and target sentences. Also, many tokens common in social media (e.g, username mentions, URLs, and emoticons) have no clear translations and annotators left them unchanged.
    \item \textbf{UBC\_HImt} \citep{jawahar-etal-2021-exploring}. They propose a dependency-free method for generating code-switched data from bilingual distributed representations and adopt a curriculum learning approach where they first fine-tune a language model on synthetic data then on gold code-switched data. 
    \item \textbf{IITP-MT} \citep{appicharla-etal-2021-iitp}. They propose an approach to create a code-switching parallel corpus from a clean parallel corpus in an unsupervised manner. Then they train a neural machine translation model on the gold corpus along with the generated synthetic code-switching parallel corpus.
    \item \textbf{UBC\_ARmt}  \citep{nagoudi-etal-2021-investigating}. They collect external parallel data from online resources and fine-tune a sequence-to-sequence transformer-based model on external data then on gold code-switched data.
    \item \textbf{CMMTOne} \citep{dowlagar-mamidi-2021-gated}. They present a gated convolutional sequence to sequence encoder and decoder models for machine translation on code-switched data. The sliding window inside the convolutional model renders it able to handle contextual words and extract rich representations.
    \item \textbf{LTRC-PreCog} \citep{gautam-etal-2021-comet}. They propose to use mBART, a pre-trained multilingual sequence-to-sequence model and fully utilize the pre-training of the model by  transliterating the roman Hindi words in the code-mixed sentences to Devanagri script.
\end{itemize}{}

In Table \ref{tab: system_comparison}, we listed the main components and strategies used by the participating systems. Most systems are based on transformer architectures, including multilingual transformers such as mT5 and mBART. It is worth noting that all teams applied deep neural networks techniques. Among the five participating teams, three of them (\textit{IITP-MT, UBC\_ARmt, CMMTOne}) trained the model from scratch while two of them (\textit{UBC\_HImt, LTRC-PreCog}) fine-tuned the models with pre-trained knowledge on monolingual data. To tackle the challenge of low-resource data, two teams (\textit{IITP-MT, LTRC-PreCog}) transliterated Hindi to Devanagari during training to fully utilize monolingual resources available in the native Devanagari script, and then back-transliterated Devanagari to Hindi to improve performance. Additionally, four teams (\textit{UBC\_HImt, IITP-MT, UBC\_ARmt, LTRC-PreCog}) leveraged external resources (e.g, parallel datasets and text processing libraries). Moreover, two teams (\textit{UBC\_HImt, IITP-MT}) applied data augmentation techniques to generate synthetic data to increase the size of the training data.

\begin{table*}[t]
\small
    \centering
    \small
    \renewcommand{\arraystretch}{1.2}
        \resizebox{\linewidth}{!}{
            \begin{tabular}{lccccc}
                \hline
                \bf System      & \bf Base Model        & \bf Training          & \bf Transliteration   & \bf Ext Resources     & \bf Data Augmentation \\
                \hline
                UBC\_HImt       & mT5                   & two-stage fine-tuning &                       & \checkmark            & \checkmark\\
                IITP-MT         & transformer           & from scratch          & \checkmark            & \checkmark            & \checkmark\\
                UBC\_ARmt       & transformer           & from scratch          &                       & \checkmark                      &\\
                CMMTOne         & seq2seq               & from scratch          &                       &                       &\\
                LTRC-PreCog     & mBART                 & fine-tuning           & \checkmark            & \checkmark            &\\
                \hline
            \end{tabular}
        }
    \caption{The comparison of the main components and strategies used by the participating systems. Ext Resources stands for external resources such as parallel datasets and text processing libraries. }
    \label{tab: system_comparison}
\end{table*}

\section{Evaluation and Results}

\subsection{Evaluation Metric}
The evaluation of the shared task was conducted through a dedicated platform, where participants can obtain immediate feedback of their submissions after uploading translations for the test data. The platform  then scores the submissions and publishes the results in a public leaderboard for each language pair direction. 

We use BLEU \citet{papineni-etal-2002-bleu} to rank participating systems. BLEU is a score for comparing a candidate translation of the text to one or more reference translations. 

\subsection{Results}
\begin{table*}[t]
\small
    \centering
    \small
    \renewcommand{\arraystretch}{1.2}
        \resizebox{\linewidth}{!}{
            \begin{tabular}{lcccccc}
                \hline
                \bf System      & \bf Eng $\rightarrow$ Hinglish    & \bf Spanglish $\rightarrow$ Eng   & \bf Eng $\rightarrow$ Spanglish   & \bf MSAEA $\rightarrow$ Eng     & \bf Eng $\rightarrow$ MSAEA \\
                \hline
                \multicolumn{6}{l}{\bf \textit{Baselines}}\\
                \hline
                baseline1       & 11.00     & 33.97     & 55.05     & 1.55      & 36.11\\
                Echo            & 6.84      & 43.86     & 65.46     & 1.80      & 40.83\\
                \hline
                \multicolumn{6}{l}{\bf \textit{Participating Systems}}\\
                \hline
                UBC\_HImt       & \bf 12.67 &           &           &           & \\
                IITP-MT         & 10.09     &           &           &           & \\
                UBC\_ARmt       &           &           &           & \bf 25.72 & \\
                CMMTOne         & 2.58      &           &           &           & \\
                LTRC-PreCog     & 12.22     &           &           &           & \\
                \hline
            \end{tabular}
        }
    \caption{The results of the participating systems in all language pairs. The scores are based on the average BLEU metric. The highlighted systems are the best scores for each language pair. }
    \label{tab: results}
\end{table*}
As listed in Table \ref{tab: results} 5 teams participated in this shared task.  4 teams submitted systems for  Eng$\rightarrow$Hinglish language directions, and  one team submitted systems for the MSAEA$\rightarrow$Eng language direction. 

\paragraph{Baselines}
For the supervised task (i.e, Eng$\rightarrow$Hinglish), \textit{baseline1} achieves a BLEU score of 11.00\%. It outperformed the \textit{Echo} baseline by 4.16\%, indicating that the multilingual model has acquired some useful knowledge about code-switching during pretraining. This is aligned with findings in literature about the usefulness of pretraining transformer models with in domain data \citep{doddapaneni2021primer}. For the unsupervised tasks, we simply use the model to make predictions based on the pre-trained knowledge it has learned on monolingual data. Surprisingly, on all four unsupervised tasks, the results from \textit{baseline1} are lower than the ones from \textit{Echo}, suggesting that the pre-trained knowledge from the monolingual data cannot be effectively adapted to the code-switched data.

\paragraph{Participating systems}
Table \ref{tab: results} shows the BLEU score results for all the teams/systems. For Eng$\rightarrow$Hinglish, two teams (\textit{UBC\_HImt, LTRC-PreCog}) outperform the baseline. The best performance achieved was from \textit{UBC\_HImt} with a BLEU score of 12.67\%, outperforming the baseline by 1.67\%. For the task of MSAEA$\rightarrow$Eng, the only one submission received was from \textit{UBC\_ARmt} with a BLEU Score of 25.72\%, outperforming the baseline by 23.92\%.

\section{Analysis}
\begin{table*}[t!]
\small
\resizebox{\linewidth}{!}{
\begin{tabular}[t]{@{}llll@{}}
\toprule
\textbf{\#} 
    & \textbf{Sentence} 
    & \textbf{Gold Standard/Reference Translation} 
    & \textbf{System Prediction} 
\\\midrule

1 & 
\begin{tabular}[t]{@{}p{0.34\linewidth}@{}}
    Yes you definitely should
\end{tabular} & 
\begin{tabular}[t]{@{}p{0.34\linewidth}@{}}
    haan tumhe definitely dekhna chahiye
\end{tabular} & 
\begin{tabular}[t]{@{}p{0.34\linewidth}@{}}
    haan, \textbf{{[}jarur aapako{]}}
\end{tabular}
\\\midrule

2 & 
\begin{tabular}[t]{@{}p{0.34\linewidth}@{}}
    It's about a mute cleaner and she works at a secret government lab 
\end{tabular} & 
\begin{tabular}[t]{@{}p{0.34\linewidth}@{}}
    ye ek mute cleaner ke baare mein hai jo secret government lab mein work karti hai
\end{tabular} & 
\begin{tabular}[t]{@{}p{0.34\linewidth}@{}}
    yah ek mute cleaner ke baare mein hai aur \textbf{{[}wo{]}} secret government lab par kaam karta hai
\end{tabular}
\\\midrule

3 & 
\begin{tabular}[t]{@{}p{0.34\linewidth}@{}}
    I wasn't sure about those two to start out with, they just didn't seem like they'd be good together
\end{tabular} &
\begin{tabular}[t]{@{}p{0.34\linewidth}@{}}
    mujhe un donon ke baare mein nishchit nahin tha ki ve shuroo kar sakate hain, unhen aisa nahin lagata tha ki ve ek saath achchhe honge
\end{tabular} & 
\begin{tabular}[t]{@{}p{0.34\linewidth}@{}}
    mujhe un dono ke baare mein \textbf{{[}start out mein sure nahi tha{]}}, unhe aisa \textbf{{[}lagta hai wo acche sath mein rehne nahin lagte hain{]}}
\end{tabular}
\\\midrule

4 & 
\begin{tabular}[t]{@{}p{0.34\linewidth}@{}}
    I mean not rated well
\end{tabular} & 
\begin{tabular}[t]{@{}p{0.34\linewidth}@{}}
    mujhe lagta hai ki ise thik se rate nahi kiya gaya
\end{tabular} & 
\begin{tabular}[t]{@{}p{0.34\linewidth}@{}}
    mera mathalab \textbf{{[}yah nahem tha ki{]}}
\end{tabular} 
\\\midrule

5 & 
\begin{tabular}[t]{@{}p{0.34\linewidth}@{}}
    The shape of water is a great movie
\end{tabular} & 
\begin{tabular}[t]{@{}p{0.34\linewidth}@{}}
    The shape of water ek bahut achchi movie hai
\end{tabular} & 
\begin{tabular}[t]{@{}p{0.34\linewidth}@{}}
    \textbf{{[}shailee{]}} kee ek mahaan philm hai
\end{tabular} 
\\\bottomrule
\end{tabular}
}
\caption{
Poor translation Examples for the task of Eng$\rightarrow$Hinglish. Errors in translations from participating systems are highlighted in \textbf{bold} in brackets (e.g, \textbf{{[}shailee{]}}). 
Reference translations are provided for comparison. 
}
\label{tab:error_examples}
\end{table*}

Although most of the scores reported by the participants outperformed the baselines for both Eng$\rightarrow$Hinglish and MSAEA$\rightarrow$Eng,  we notice that the BLEU metric takes overlapped tokens into consideration, which may potentially result in a higher score. Additionally, the overall results are arguably lower compared to results achieved by the state-of-the-art systems on monolingual data for these language pairs, i.e. in the absence of code-switching. We conduct error analysis to understand and expose the bottleneck of machine translation on code-switched data.

\subsection{Evaluation Metrics}
In Table \ref{tab: results}, it is worth noting that the \textit{Echo} system, with inputs as outputs, could outperform the transformer-based \textit{baseline1} system in four out of five language translation directions. We suspect the high scores from the \textit{Echo} system are because of the high overlap between the training and testing data. Therefore, to minimize such impact, we normalize the text by removing the overlapping tokens from the data, including username mentions, emoji expressions, and URLs. The results of BLEU and normalized BLEU scores on each system for the task of Eng$\rightarrow$Hinglish are shown in Table \ref{tab: bleu_comparision} where we note significant drops in performance ranging from 0.43\% to 6.13 BLEU score. After normalizing the text, the \textit{Echo} system can only reach 0.71 BLEU points, which is far lower than the performance of \textit{baseline1}. Although the performances of participating systems also decrease drastically, some can still outperform the \textit{baseline1}.
\begin{table}[ht]
\small
    \centering
    \small
    \renewcommand{\arraystretch}{1.2}
        \resizebox{0.8\linewidth}{!}{
            \begin{tabular}{llll}
                \hline
                \bf System      & \bf BLEU & \bf BLEU\_norm & \bf Drop \\
                \hline
                \multicolumn{4}{l}{\bf \textit{Baselines}} \\
                \hline
                Baseline1       & 11.00     & 6.05      & 4.95 \\
                Echo            & 6.84      & 0.71      & \bf 6.13\\
                \hline
                \multicolumn{4}{l}{\bf \textit{Participating Systems}} \\
                \hline
                UBC\_HImt       & \bf 12.67 & \bf 7.58  & 5.09 \\
                IITP-MT         & 10.09     & 6.33      & 3.76 \\
                CMMTOne         & 2.58      & 2.15      & 0.43 \\
                LTRC-PreCog     & 12.22     & 6.70      & 5.52 \\
                \hline
            \end{tabular}
        }
    \caption{The BLEU and the normalized BLEU scores on the task of Eng$\rightarrow$Hinglish. The largest number for each column is in \textbf{bold}.}
    \label{tab: bleu_comparision}
\end{table}

\subsection{Error Analysis}
Although most of the scores reported by the participants outperform the baseline, there are many mistakes made by the models. Thus, we manually inspect the predictions from participating systems. Table \ref{tab:error_examples} shows some cases where the participating systems produce low quality translations for the task of Eng$\rightarrow$Hinglish.

Based on our findings, there are two main errors among the predictions: \textit{grammar errors} and \textit{semantic errors}. For the \textit{grammar errors}, we find that many predicted sentences have incorrect word/phrase orders. Although linguistic theories of code-switching argue that code-switching does not violate syntactic constraints of the languages involved \citep{poplack1980sometimes, poplack2013sometimes}, it may still impact models ability to correctly capture grammatical structures from the involved languages. Moreover, we observe that some translations may have extra punctuation or incorrect pronouns. Gender translation errors are also common in MT systems for monolingual data \citep{stanovsky-etal-2019-evaluating, saunders-byrne-2020-reducing}. For the \textit{semantic errors}, it is worth noting that many sentences can only be translated partially. Understanding the semantics of code-switching is difficult as the model has to learn the word representations and dependencies for all involved languages. This may result in incomplete or irrelevant translations.

\section{Conclusion}
This paper summarizes the insights gained from creating a task on MT for code-switched language. Given a monolingual/code-switched sentence, the goal is to translate it into a code-switched/monolingual sentence. We introduced a machine translation dataset focused on code-switched social media text for three language pairs: English-Hinglish, English-Spanglish, and English-Modern Standard Arabic-Egyptian Arabic with one or two directions for each pair. We received 12 submissions from 5 teams, 4 of them submitted to English$\rightarrow$Hinglish and 1 of them submitted to Modern Standard Arabic-Egyptian Arabic$\rightarrow$English. While the participating systems are different, we observed that there is a strong trend favoring the use of transformer-based models (e.g, mBART) as a key ingredient in the proposed architectures. The best performances achieved were 12.67\% BLEU score for English$\rightarrow$Hinglish and 25.72\% BLEU score for Modern Standard Arabic-Egyptian Arabic$\rightarrow$English. Compared to monolingual formal text, the reported BLEU scores are significantly lower, which highlights the difficulty of both, the linguistic properties in code-switched data, and the noise attributed to the social media genre. More investigations need to be carried out to disentangle the effects of each factor. However, this shared task demonstrates that performing machine translation on code-switched data is a new horizon of interest in the NLP community and more work needs to be done to bring performance on par to monolingual systems. It is also likely that advances in MT for code-switched data will lead to groundbreaking contributions to the general field of MT.

\section*{Acknowledgements}
As organizers of the CALCS 2021 shared task, we gratefully acknowledge all the teams for their interest and participation. We also would like to thank the Alan W. Black, Sunayana Sitaram, Victor Soto and Emre Yilmaz for their invaluable collaboration in organizing this shared task.

\bibliography{anthology,custom}
\bibliographystyle{acl_natbib}

\end{document}